# Can you tell?
SSNet - a Biologically-inspired Neural Network Framework for Sentiment Classifiers

Apostol Vassilev
Munawar Hasan
Honglan Jin

**National Institute of Standards and Technology**

{apostol.vassilev, munawar.hasan, honglan.jin}@nist.gov

**Abstract**

When people try to understand nuanced language they typically process multiple input sensor modalities to complete this cognitive task. It turns out the human brain has even a specialized neuron formation, called sagittal stratum, to help us understand sarcasm. We use this biological formation as the inspiration for designing a neural network architecture that combines predictions of different models on the same text to construct robust, accurate and computationally efficient classifiers for sentiment analysis and study several different realizations. Among them, we propose a systematic new approach to combining multiple predictions based on a dedicated neural network and develop mathematical analysis of it along with state-of-the-art experimental results. We also propose a heuristic-hybrid technique for combining models and back it up with experimental results on a representative benchmark dataset and comparisons to other methods[1] to show the advantages of the new approaches.

# Introduction

Applications of deep learning to natural language processing represent attempts to automate a highly-sophisticated human capability to read and understand text and even generate meaningful compositions. Language is the product of human evolution over a very long period of time. Scientists now think that language and the closely related ability to generate and convey thoughts are unique human traits that set us apart from all other living creatures. Modern science describes two connected but independent systems related to language: inner thought generation and sensor modalities to express or take them in for processing [9]. For example, human sensory modalities are speaking, reading, writing, etc. This allows homo sapiens to express an infinite amount of meaning using only a finite set of symbols. e.g. the 26 letters in the English language. The result is a very powerful combination that has resulted in the vast amount of knowledge and information amassed in the form of written text today.

Over the course of the long evolutionary development and especially in the modern era, the sapiens have mastered the ability to generate and convey sophisticated and nuanced thoughts. Consequently, the texts deep learning is tasked with processing, known as natural language processing (NLP), range from the simple ones that say what they mean to those that say one thing but mean another. An example of the latter is sarcasm. To convey or comprehend sarcasm the sapiens typically invoke more than one sensory modality, e.g. combining speech with gestures or facial expressions, or adding nuances to the speech with particular voice tonalities. In written text, comprehending sarcasm amounts to what colloquially is known as reading between the lines.

With the emergence of A.M. Turing's seminal paper [38], the research in NLP kicked off. Initially, it revolved around handwritten rules, later obsoleted by statistical analysis. Until the advent of deep learning, the decision tree based parsing techniques [26, 27] were considered as the state of the art methods and linguistic performance was measured on the basis of the *Turing Test* [38].

Fast forward to present day, deep learning and the enormous increase in available computational power allowed researchers to revisit the previously defined as computationally intensive family of recurrent neural networks [19, 36] and produced several groundbreaking NLP results [8, 20, 40, 42]. There are also numerous other application-specific architectures [5, 10, 13] for NLP problems. We refer the reader to a recent comprehensive survey [29] for a detailed review of a large number of deep learning models for text classification developed in recent years and a discussion of their technical contributions, similarities, and strengths. This survey also provides a large list of popular datasets widely used for text classification along with a quantitative analysis of the performance of different deep learning models on popular benchmarks. Still, it is a publicly-held secret that the algorithms used in machine learning, including the most advanced, are limited in the sense that all of them fail to capture *all* information contained in the data they process. Recent research even points to a systemic problem known as underspecification [12, 17].

Having been faced with this reality, we asked ourselves the question: How do humans do it? How do they process ambiguous text? We all know that our human sensor abilities are limited: our vision, our hearing, our attention span, our memory are all limited. How do we then perform so well given all these limitations?

Through the evolution of their brain, sapiens have acquired a polygonal crossroad of associational fibers called sagittal stratum (SS), cf. Figure 1[2], to cope with this complexity. Researchers have reported [34] that the bundle of nerve fibers that comprises the SS and connects several regions of the brain that help with processing of information enables people to understand sarcasm through sensory modalities – both visual information, like facial expressions, and sounds, like tone of voice. Moreover, researchers have shown that the patients who had the most difficulty with comprehending sarcasm also tended to have more extensive damage in the right SS. In other words, the ability to understand sophisticated language nuances is dependent on the ability of the human brain to successfully take in and combine several different types of sensory modalities.

---

[2]Reprinted from [6] with permission by Springer Nature, order #4841991468054.

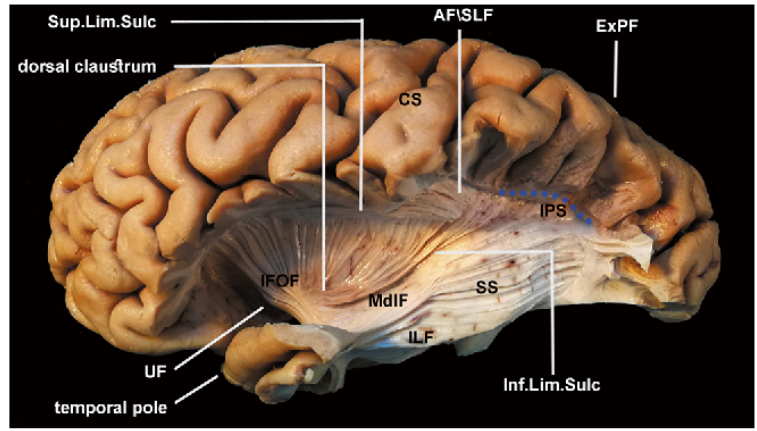

**Figure 1. This image shows the sagittal stratum (SS).** The SS is situated deep on the lateral surface of the brain hemisphere, medial to the arcuate/superior longitudinal fascicle complex, and laterally to the tapetal fibers of the atrium [6]. The SS is a bundle of nerve fibers that connects many different parts of the brain and helps with processing sensory modalities (visual and sound) and thus enables people to understand nuanced language such as sarcasm.

The evolution of language and the resulting increased sophistication of expressing human thoughts has created a challenging problem for deep learning. How to capture and process the full semantics in a text is still an open problem for machine learning. This is partly manifested by the facts that first, there are many different ways of encoding the semantics in a text, ranging from simple encoding relying on treating words as atomic units represented by their rank in a vocabulary [3], to using word embeddings or distributed representation of words [28], to using sentence embeddings and even complete language models [14, 18, 37]; second, there is no established dominant neural network type capable of successfully tackling natural language processing in most of its useful for practice applications to the extent required by each specific application domain.

Based on this observation, we explored the extent to which it is possible to utilize a simple encoding of semantics in a text and define an optimal neural network for that encoding [39] for sentiment analysis. Our study showed that although each of these encoding types and corresponding neural network architecture may yield good results, they are still limited in accuracy and robustness when taken by themselves.

The main thrust of NLP research is based on the idea of developing computationally intensive neural network architectures intended to produce better results in the form of accuracy on representative benchmark datasets. In contrast, the research on simulating the decision making capabilities of our brain related to perception or past experiences with machine learning has lagged. Thus, the computed probability of any linguistic sample predicted by any individual model is not grounded in a state or function of a biological mind. As we mentioned above, the anatomy of the human brain allows processing of multiple input sensor modalities to make a decision. Inspired by this, this paper seeks to establish a novel approach to sentiment analysis for NLP.

The primary goal of this paper is to explore the problem from a different perspective and to study ways to combine different types of encoding intended to capture better the semantics in a text along with a corresponding neural network architecture inspired by the SS in the human brain. To do this, we introduce a new architecture for neural network for sentiment analysis and draw on the experiences from using it with several different types of word encoding in order to achieve performance better than that of each individual participating encoding. The main contribution of this paper is the design of the biologically-inspired framework for neural networks for sentiment analysis in Section 2 and the analysis of the combiner based on a neural network in Section 2.1.

The authors would like to emphasize that this paper does not try to improve the metrics achieved by aforementioned papers but presents an approach to simulate certain decision making scenarios in the human brain. Any model referenced in this section can be used as a plug and play module in our framework.

# 1 Limitations of existing standalone NLP approaches to machine learning

As indicated above, there are multiple different types of encoding of semantics in text, each of varying complexity and suitability for purpose. The polarity-weighted multi-hot encoding [39], when combined with appropriately chosen neural network, is generic yet powerful for capturing the semantics of movie reviews for sentiment analysis. Even though the overall accuracy reported in [39] is high, the approach quickly reaches a ceiling should higher prediction accuracy be required by some application domains.

Encoding based on word embeddings or distributed representation of words [28] is widely used. For example, the approach in [32] has been influential in establishing semantic similarities between the words for a given corpus, by projecting each word of the corpus to a high dimensional vector space. While the dimension of the vector space itself becomes a hyperparameter to tweak around, the vectors can be further processed or utilized using a recurrent neural network (RNN). When tackling NLP problems, a variant of RNN, namely the long short term memory (LSTM) variant and its bidirectional version (BLSTM) are known to perform better than other neural networks. Through our experiments on various datasets [1, 21], we found that certain vocabulary provides deeper semantics to the sentence from the corpus based on the receiver's perception and context. In such situations, the idea of attention [2, 16, 25] plays an important role and provides the network with an additional parameter called the context vector, which can make convergence slower, but the model overall is robust. It is also possible to use a learnable word embedding, where the first layer of the neural network architecture is the embedding followed by one or more RNN's.

Although intuitively one may think that word embeddings should help to increase the accuracy of the model to any desirable level because word embeddings do capture the semantics contained in the text in a way that mimics how people perceive language structure, the available empirical test evidence in terms of reported accuracy rates is inconclusive. Our own experiments with word embeddings used by themselves revealed an accuracy ceiling similar to that of the polarity-weighted multi-hot encoding. Attempts to utilize sentence embeddings have been even less successful [11].

Recently, pretrained language models have gained popularity because of their improved performance on general NLP tasks [14]. These models are trained on large corpora, and their applications to specific NLP applications typically involves some transfer learning on the corpus of interest. However, they too have limitations, which we will discuss in more detail in Section 3.2.3.

All these different types of encoding can be challenged further by varying style of writing or level of mastering the language. Examples of the former are nuanced language such as sarcasm. Some reviewers choose to write a negative review using many positive words yet an experienced reader can sense the overall negative sentiment conveyed between the lines while the polarity-weighted multi-hot encoding [39] and word embeddings [32] may struggle with it. Examples of the latter are primitive use of the language by non-native speakers resulting in sentences with broken syntax and inappropriate terminology. Other difficult cases are reviews that contain a lot of narrative about the plot of the movie but very little of the reviewer's opinion about how she feels about the movie. Yet another problematic class are movie reviews that rate a movie excellent for one audience, e.g. children, but not good for another, e.g. adults. Careful analysis of the data in [1, 21] reveals examples of all these kinds of reviews, often confusing models based on the encodings described here. Such complications represent significant challenges to each of these types of encoding when used by themselves, no matter the power of the neural network.

This observation raises a question: if one is interested in obtaining a more robust and versatile representation of the semantics of text would an approach that combines different types of encoding yield a better result than attempting to just improve each of them within their envelopes?

## 2 Sagittal stratum-inspired neural network

We now turn to the design of a neural network that aims to simulate the way SS in the human brain operates. Recall that the SS brings information from different parts of the brain, each responsible for processing different input sensory modalities, to enable a higher order of cognition, such as comprehension of sarcasm. Our context here is NLP and one way to map the functioning of the SS to it is to consider combining different representations of the semantic content of a text. To do this, one first has to pick the types of representations of the text. Because we aim at computing different perspectives on the same text, it is natural to seek representations that are independent. For example, the polarity-weighted multi-hot encoding [39] is based on the bag-of-words model of the language and has nothing in common with word embeddings that rely on language structure [32] or the transformer language model [14]. But if independence of representation is adopted, how does one combine the models computed from each of them?

Unlike image processing where each model is computed over the pixel grid of the image, in NLP there is no common basis onto which to map and combine the different models. Instead, we again use a hint from how the human brain performs some cognitive tasks. When a person hears another person utter a phrase, to comprehend what the speaker is trying to convey the brain of the listener first processes the words in the phrase, then the listener assesses if the speaker rolled her eyes, for example, when uttering the words, to decide if she spoke sarcastically. The brain of the listener combines these two assessments with the help of the SS to arrive at a final conclusion if the speaker spoke sarcastically or not. This suggest we can combine the resulting assessments from each model on a particular review $\tau$, e.g., the probability of classifying it as positive or negative, to decide on the final classification.

The neural networks based on the different language models are trained on the same corpus. In the case of transformer models, which are pre-trained on extremely large corpora, they are transfer trained on the same corpus as the remaining models to ensure proper adaptation to the problem at stake. The trained models are saved and used to compute predictions. For the sake of developing notation, we assume there are $K$ different models.

The predictions from the $K$ models are fed into a SS-inspired classifier. Based on our understanding for how SS works in the human brain to enable interpretation of language that can only be resolved correctly when multiple sensor modalities are used together, we construct the network shown in Figure 2 and experiment with several different mechanisms F for combining the models.

In general the computed probabilities for review classification from the participating models are combined in a way that favors identifying the most probable case, which is analogous to the way humans assess multiple sensor modalities in order to process ambiguous speech and deduce the most plausible interpretation. Our task is to combine the predictor models taking into account that they represent different views on the semantics in the same text. Recall from the observation in Section 2 that the only meaningful way to combine the models is through the probability assessment each of them produces for a given review and in turn the entire corpus. While the ensemble techniques [30] have been known for good performance in computer vision related tasks [23, 31, 33, 35], the same is not true for natural language processing based problems. This analogy can also be backed by the fact that models using different encodings have different latent space and hence merging such latent spaces may not produce an optimal solution due to the varying rate of convergence of individual models. But the major issue is the projection of one model's latent space onto another. Due to different encodings, such projections may produce inconsistent coalesced models.

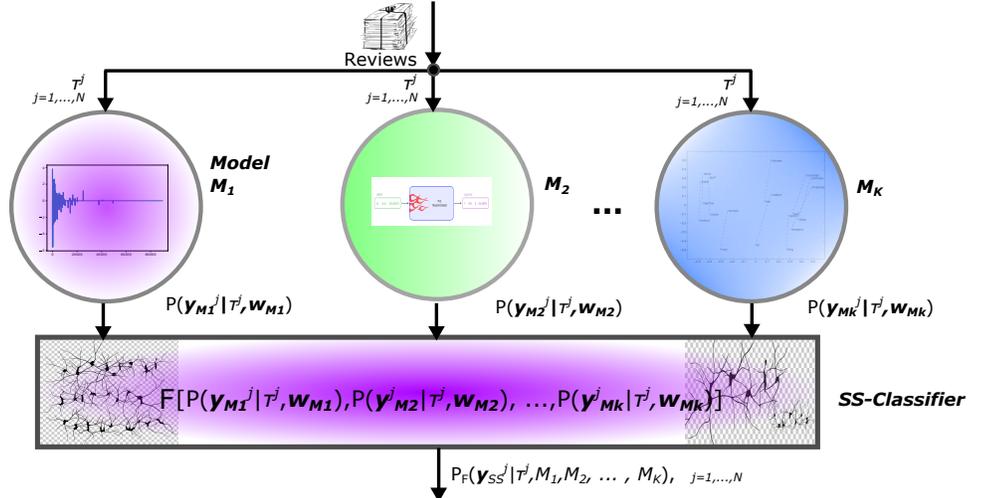

**Figure 2. The SS classifier**. Here F is some appropriately defined function that combines the input from the participating models, and $\tau^j$, $\forall j \in \{1, ..., N\}$, are the input text samples. $M_1$ to $M_K$ are the $K$ participating models, and $\mathrm{P}(\mathbf{y}^j_{M_i}|\tau^j, \mathbf{w}_{M_i})$ is the probability computed by $M_i$, $\forall i \in \{1, ..., K\}$, for each $\tau^j$. Finally, $\mathrm{P}_\mathrm{F}(\mathbf{y}^j_{SS}|\tau^j, M_1, ..., M_K)$, is the resulting probability for $\tau^j$ computed by the combiner F.

One other potential concern here is that the models may be strongly correlated as they are trying to predict the same thing. We compensate for this by using fundamentally different language models. Each of the models has different limitations that are challenged by the text of the different reviews in the dataset, yielding different predictions on some of them. In addition, we apply random shuffling of the training dataset on each training epoch. These two factors alleviate this potential concern to a large extent, as confirmed by the computational results in Section 3. Thus, our approach is different than the classic leave-one-out training approach in [4, 24] and better suited for the current state-of-the-art NLP models and their applications on large datasets of interest.

In principle, there are two potential approaches to combining the models: systematic and heuristic. A systematic approach avoids the use of heuristics in deciding how to combine the models. Instead, one states assumptions on the properties of an individual model and then builds a general rule to combine the models. Following this principle, we introduce a neural network combiner in Section 2.1 and analyze its properties. This approach delivers state-of-the-art accuracy results on the benchmark corpus of interest, which we provide in Section 3. For a baseline comparison we consider the systematic Bayesian combination rules in [22] (**Sum**, **Majority Vote**, **Average**, **Max**). A brief discription of this is given in Section 2.2.

We also propose a hybrid heuristic-systematic technique for combining models in Section 2.3, which makes use of the combination rules in [22] but with our heuristic way of choosing what and when to combine. This hybrid technique shows performance characteristics that are pretty close to the leading neural network combiner and outperforms the classic Bayesian rule combiners - see Section 3.

## 2.1 A neural network combiner

Here we seek to define a predictor F inspired by human biology and in terms of a neural network consisting of a single dense layer and a sigmoid. The neural network computes the weights $\{w_i\}_{i=1}^{K}$ for combining the participating models $M_1, ..., M_K$. Let $\mathbf{y}_i(\tau^j)$ be the probability estimate computed by the $i$-th model on the $j$-th text, denoted as $P(y_{M_i}|\tau^j, w_{M_i})$ in Figure 2.

We define the combined predictor as a combination of $K > 1$ individual predictors:

$$\mathbf{y}(\tau) = \sum_{i=1}^{K} w_i \mathbf{y}_i(\tau), \ \forall \tau \in \mathbb{D}, \tag{1}$$

$$w_i \geq 0, \ \forall i. \tag{2}$$

In the case of a corpus labeled for binary classification, a binary function $\mathbf{u}(\tau)$ is defined by the label assigned to each text $\tau$ in $\mathbb{D}$. Given the two classes, $\mathbf{I}^{(0)}$ and $\mathbf{I}^{(1)}$, $\mathbf{y}_i(\tau)$ is the predicted probability for the text $\tau$ to belong to $\mathbf{I}^{(1)}$. The real-valued functions $\mathbf{y}_i$ with range the interval $[0, 1]$ have values that correspond to the probability of being assigned to $\mathbf{I}^{(1)}$. Because of 1 and 2, the range of $\mathbf{y}(\tau)$ may exceed the unit interval, so typically one assigns the class by subjecting $\sigma(\mathbf{y}(\tau))$ to a threshold test with some value $t \in (0, 1)$ so that

$$\tau \in \begin{cases} \mathbf{I}^{(1)} & \text{, if } \sigma(\mathbf{y}(\tau)) \geq t, \\ \mathbf{I}^{(0)} & \text{, otherwise.} \end{cases} \tag{3}$$

Here $\sigma(x)$ is the sigmoid function given by

$$\sigma(x) = \frac{1}{1 + e^{-x}}.$$

Because $\mathbf{y}_i(\tau)$ are with ranges shifted with respect to the domain of the sigmoid, to get accurate classification one needs to shift the range of $\mathbf{y}$ to the left so it is centered with respect to zero, i.e.,

$$\sigma_b(\mathbf{y}(\tau)) = \sigma(\mathbf{y}(\tau) - b), \tag{4}$$

for some $b > 0$. Here, we assume that each predictor $\mathbf{y}_i$ is decent, i.e., it produces labels that are reasonably close to those produced by $\mathbf{u}$ over the entire $\mathbb{D}$. Mathematically, this means we assume $||\mathbf{u} - \mathbf{y}_i||$ is small, compared to $||\mathbf{u}||$. Here,

$$||\mathbf{f}||^2 = \sum_{\tau \in \mathbb{D}} \mathbf{f}^2(\tau), \tag{5}$$

where $\mathbf{f}$ is a binary function defined over $\mathbb{D}$.

For the case of a real-valued function $\mathbf{y}_i(\tau)$ we define

$$||\mathbf{y}(\tau)||_t^2 = \sum_{\tau \in \mathbb{D}} \mathbf{I}_t(\mathbf{y}(\tau))^2, \tag{6}$$

where $\mathbf{I}_t$ is the assigned class for $\mathbf{y}(\tau)$ with respect to the threshold $t$ according to 3. Similarly, we define

$$||\mathbf{y}(\tau) - \mathbf{z}(\tau)||_t^2 = \sum_{\tau \in \mathbb{D}} (\mathbf{I_t}(\mathbf{y}(\tau)) - \mathbf{I_t}(\mathbf{z}(\tau)))^2. \quad (7)$$

Note that for a binary function $\mathbf{f}$ over $\mathbb{D}$, $||\mathbf{f}|| = ||\mathbf{f}||_t$. Note also that the definitions 5 and 6 imply that $||\mathbf{u}||$ is large. If $N$ is the cardinality of $\mathbb{D}$, then in fact $||\mathbf{u}||$ is close to $N/2$. Otherwise, $\mathbf{u}$ would be statistically biased to one of the classes. Also, $N$ is large for problems of interest, otherwise the data can be processed by humans. Related to $N$, the number $K$ of individual predictors is small, typically up to a half a dozen.

### 2.1.1 Analysis of the neural network combiner

In this section we consider the question if the constraint 2 is sufficient for computing reasonable weights to use in combining the models. In the literature, people often impose the additional constraint

$$\sum_{i=1}^{K} w_i = 1. \quad (8)$$

However, we are not aware of a similar constraint imposed by the SS or another region of the brain when handling the different input modalities, based on our review of the literature. Humans seem to be making decisions when they consider all input modalities together in the proper context, not by discarding some of the input modalities upfront. So, it is reasonable to want the weights of the different inputs to our system be of the same order, i.e., there is not one input whose weight dominates the others, effectively reducing the mix to a single modality. All that without imposing artificial constraints, e.g., (8), that do not have clear backing in the biological foundations of our inspiration. Moreover, after performing extensive computations in the laboratory, we observed that imposing 8 makes computing the optimal weights much more intense and difficult without any gains in the accuracy of the combined predictor. This led us to examine the need for 8. We argue that the additional constraint 8 is **not** necessary. To see this, let us consider the approximation error of the predictor $\mathbf{y}$ defined as $||\mathbf{u} - \mathbf{y}||_t$. Let us denote $W = \sum_{i=1}^{K} w_i$. Let

$$\hat{\mathbf{y}}(\tau) = \frac{1}{W}\mathbf{y}(\tau) = \sum_{i=1}^{K} \frac{w_i}{W}\mathbf{y}_i(\tau) \quad (9)$$

be the interpolation predictor constructed as a liner combination of $\mathbf{y}_i$ with coefficients that sum up to one. If the individual predictors are good then the interpolation predictor $\hat{\mathbf{y}}$ is also good, i.e. $||\mathbf{u} - \sigma_b(\hat{\mathbf{y}})||_t$ is small.

Let $\mathbb{L}_t(x)$ be a linear approximation of $\sigma(x)$ for some constant $t > 0$ such that $\mathbb{L}_t(x)$ minimizes $||\mathbb{L}_t(x) - \sigma(x)||_t$. Note that any straight line passing through the points $(\ln(\frac{t}{1-t}), t)$ and having the same slope as $\sigma'(\ln(\frac{t}{1-t}))$ satisfies $||\mathbb{L}_t(x) - \sigma(x)||_t = 0$. For example, for $t = \frac{1}{2}$ the straight line

$$\mathbb{L}_{\frac{1}{2}}(x) = \frac{1}{4}x + \frac{1}{2}$$

also satisfies $||\mathbb{L}_{\frac{1}{2}}(x) - \sigma(x)||_{\frac{1}{2}} = 0$. In practice $t = \frac{1}{2}$ is the natural choice for an unbiased binary distribution $\mathbf{u}$ and an unbiased predictor $\mathbf{y}$. Note that instead of taking the entire straight line, one may consider a piece-wise linear function but this is not necessary because the definitions 6 and 7 can handle unbounded functions.

Then,
$$||\mathbf{u} - \sigma_b(\mathbf{y})||_t = ||\mathbf{u} - W\mathbf{u} + W\mathbf{u} - \sigma_b(\mathbf{y})||_t =$$

$$||(1-W)\mathbf{u} - W(\mathbf{u} - \frac{1}{W}\sigma_b(\mathbf{y}))||_t.$$

Applying the triangle inequality, we get

$$||\mathbf{u} - \sigma_b(\mathbf{y})||_t \geq |1-W|\,||\mathbf{u}||_t - W||\mathbf{u} - \frac{1}{W}\sigma_b(\mathbf{y}))||_t. \tag{10}$$

First, consider the case $W \leq 1$. Then from inequality 10

$$||\mathbf{u} - \sigma_b(\mathbf{y})||_t \geq (1-W)||\mathbf{u}||_t - W||\mathbf{u} - \frac{1}{W}\sigma_b(\mathbf{y}))||_t.$$

Let $\mathbb{L}_t$ be as defined above and

$$\mathbb{L}_{t,b}(x) = \mathbb{L}_t(x-b). \tag{11}$$

Then,

$$||\mathbf{u} - \frac{1}{W}\sigma_b(\mathbf{y}))||_t = ||\mathbf{u} - \frac{1}{W}\mathbb{L}_{t,b}(\mathbf{y}) + \frac{1}{W}\mathbb{L}_{t,b}(\mathbf{y}) - \frac{1}{W}\sigma_b(\mathbf{y}))||_t$$

Thus,

$$||\mathbf{u} - \frac{1}{W}\sigma_b(\mathbf{y}))||_t \leq ||\mathbf{u} - \mathbb{L}_{t,b}(\hat{\mathbf{y}})||_t + \frac{1}{W}||\mathbb{L}_{t,b}(\mathbf{y}) - \sigma_b(\mathbf{y}))||_t.$$

Note that

$$\frac{1}{W}||\mathbb{L}_{t,b}(\mathbf{y}) - \sigma_b(\mathbf{y}))||_t = 0.$$

From here we get,

$$||\mathbf{u} - \sigma_b(\mathbf{y})||_t \geq (1-W)||\mathbf{u}||_t - W(||\mathbf{u} - \mathbb{L}_{t,b}(\hat{\mathbf{y}})||_t)$$

This implies that

$$W \geq \frac{||\mathbf{u}||_t - ||\mathbf{u} - \sigma_b(\mathbf{y})||_t}{||\mathbf{u}||_t + ||\mathbf{u} - \mathbb{L}_{t,b}(\hat{\mathbf{y}})||_t}.$$

Note that $||\mathbf{u} - \mathbb{L}_{t,b}(\hat{\mathbf{y}})||_t = ||\mathbf{u} - \sigma_b(\hat{\mathbf{y}})||_t$, because by construction $\mathbf{I}_t(\mathbb{L}_{t,b}(\hat{\mathbf{y}})) = \mathbf{I}_t(\sigma_b(\hat{\mathbf{y}}))$. Hence,

$$W \geq \frac{||\mathbf{u}||_t - ||\mathbf{u} - \sigma_b(\mathbf{y})||_t}{||\mathbf{u}||_t + ||\mathbf{u} - \sigma_b(\hat{\mathbf{y}})||_t}. \tag{12}$$

Note also that $||\mathbf{u} - \sigma_b(\mathbf{y})||_t$ is small, especially with respect to the size of $||\mathbf{u}||_t$. Similarly, by the definition of $\hat{\mathbf{y}}$ in 9, $||\mathbf{u} - \sigma_b(\hat{\mathbf{y}})||_t$ is small.

Next, consider the case $W > 1$. Then, inequality 10 implies

$$||\mathbf{u} - \sigma_b(\mathbf{y})||_t \geq (W-1)||\mathbf{u}||_t - W||\mathbf{u} - \frac{1}{W}\sigma_b(\mathbf{y}))||_t. \tag{13}$$

Introducing $\mathbb{L}_{t,b}(x)$ as in the previous case, we get

$$||\mathbf{u} - \frac{1}{W}\sigma_b(\mathbf{y}))||_t = ||\mathbf{u} - \frac{1}{W}\mathbb{L}_{t,b}(\mathbf{y}) + \frac{1}{W}\mathbb{L}_{t,b}(\mathbf{y}) - \frac{1}{W}\sigma_b(\mathbf{y}))||_t$$

Thus,

$$||\mathbf{u} - \frac{1}{W}\sigma_b(\mathbf{y}))||_t \leq ||\mathbf{u} - \mathbb{L}_{t,b}(\hat{\mathbf{y}})||_t + \frac{1}{W}||\mathbb{L}_{t,b}(\mathbf{y}) - \sigma_b(\mathbf{y})||_t.$$

As we observed above, $W^{-1}||\mathbb{L}_{t,b}(\mathbf{y}) - \sigma_b(\mathbf{y}))||_t = 0$ and $||\mathbf{u} - \mathbb{L}_{t,b}(\hat{\mathbf{y}})||_t = ||\mathbf{u} - \sigma_b(\hat{\mathbf{y}})||_t$. Hence,

$$||\mathbf{u} - \frac{1}{W}\sigma_b(\mathbf{y}))||_t \leq ||\mathbf{u} - \sigma_b(\hat{\mathbf{y}})||_t.$$

Substituting this into inequality 13 gives

$$||\mathbf{u} - \sigma_b(\mathbf{y})||_t \geq (W-1)||\mathbf{u}||_t - W||\mathbf{u} - \sigma_b(\hat{\mathbf{y}})||_t.$$

From here we get

$$||\mathbf{u} - \sigma_b(\mathbf{y})||_t + ||\mathbf{u}||_t \geq W(||\mathbf{u}||_t - ||\mathbf{u} - \sigma_b(\hat{\mathbf{y}})||_t).$$

Note that $||\mathbf{u} - \sigma_b(\mathbf{y})||_t$ and $||\mathbf{u} - \sigma_b(\hat{\mathbf{y}})||_t$ are small relative to $||\mathbf{u}||_t$, which in turn means that $||\mathbf{u}||_t - ||\mathbf{u} - \sigma_b(\hat{\mathbf{y}})||_t > 0$ and not too far from $||\mathbf{u}||_t$. This implies that

$$W \leq \frac{||\mathbf{u}||_t + ||\mathbf{u} - \sigma_b(\mathbf{y})||_t}{||\mathbf{u}||_t - ||\mathbf{u} - \sigma_b(\hat{\mathbf{y}})||_t}. \tag{14}$$

Thus, we have proved the following theorem.

**Theorem.** *Let $\mathbf{u}$ be a binary function over $\mathbb{D}$ and $\mathbf{y}$ be the combined predictor 1 that approximates it. Let $\hat{\mathbf{y}}$ be the interpolation predictor 9. Let $t \in (0, 1)$ be the threshold 3. Then, $W = \sum_{i=1}^{K} w_i$ satisfies:*

$$\frac{||\mathbf{u}||_t - ||\mathbf{u} - \sigma_b(\mathbf{y})||_t}{||\mathbf{u}||_t + ||\mathbf{u} - \sigma_b(\hat{\mathbf{y}})||_t} \leq W \leq \frac{||\mathbf{u}||_t + ||\mathbf{u} - \sigma_b(\mathbf{y})||_t}{||\mathbf{u}||_t - ||\mathbf{u} - \sigma_b(\hat{\mathbf{y}})||_t}. \tag{15}$$

*Proof.* Combine 12 and 14 to obtain 15. □

**Note:** *The theorem justifies our approach of utilizing only constraints that reflect the natural mechanisms of the SS by guaranteeing balanced and bound weights for combining the predictors, without the artificial interpolating constraint* (8).

**Note:** *The result of the above theorem is general and applies to binary classification problems involving predictors $\mathbf{y}_i$ in other application domains, beyond NLP sentiment analysis. This result can also be extended to multi-class classifications based on the well-known conversion of any multi-class classification problem to a binary classification problem using the* **one-vs-rest** *method.*

## 2.2 Bayesian decision rule combiners

Here we consider some of the Bayesian rule combiners in [22], in particular the **Sum**, **Average**, **Majority Vote**, and **Max** rules. Kittler at. al. [22] present a systematic approach to deriving the rules but acknowledge that some of the assumptions used to develop their combination rules are too restrictive for many practical applications. We present computational results from applying these rules in Section 3

## 2.3 A heuristic hybrid combiner

The idea here is to select the best model, e.g., the one with highest accuracy on the train dataset, and use it as a base model. Then use the prediction of the base model as the prediction of the combiner unless its confidence level drops below a predefined threshold value $\theta$. In that case, use instead the predictions from the other auxiliary models, and combine them using a Bayesian decision rule from [22], e.g., **Sum**, **Majority Vote**, **Average**, and **Max**.

Generally speaking, in cases where the comparison between the candidate models is inconclusive, the choice of the base and auxiliary models may be approached analogously to how humans interpret ambiguous voice: do they trust more the deciphering of the words or the evaluation of the facial expression of the speaker to decide what they mean? Some people may choose to weigh the words heavier than the voice in a given circumstance, others may opt the other way around. However, it is always important to be aware of the limitations the models may have in the context of the potential application.

# 3 Computational results

In this section, we begin with an overview and performance of the individual components of the architecture. The individual components are neural network models that are built with the same corpus. Then we present the computational results obtained with our proposed architecture that combines the results from the above components.

To address the issue of accuracy fluctuation due to the stochastic nature in neural network models, we use 5-folds cross-validation in reporting individual model performance. For our proposed SSNet combiner, we reuse the 5 predictions generated from 5 folds in each participant model as input to train the combiner. The goal is to control the performance variation brought in by randomness from the models, so we use 5 different sets of inputs to train SSNet 5 times, all with the same architecture. We go even further by also training the combiner multiple times using the same input. For example, for each of the 5 predictions, we repeat the training $k$ times, and, therefore, our total number of combiner training rounds is $5k$. In each training round we use the entire 25 000 test reviews to get a test accuracy. Then we use the mean average of accuracy computed over the $5k$ results as our final score for SSNet. We choose $k = 30$ in our experiments.

All experiments reported here were performed with our research code (available at https://github.com/usnistgov/STVM_NLP_Research) developed in Python 3.0 with the TensorFlow 2.1.0 [15] library. The models were trained on a professional Graphics Processing Unit (GPU) cluster having 8 NVIDIA Tesla V100 (32 GB each). The inference was carried on a 2015 MacBook Pro with 2.5 GHz Intel Core i7 and 16 GB RAM *without* Graphics Processing Unit (GPU) acceleration.

## 3.1 Datasets

When evaluating the performance of machine learning models, one has to keep in mind that the results depend heavily not only on the model but the dataset it was trained on. The bigger and better the training dataset, the better the results. In practice, one needs to analyze the training dataset and improve its quality before commencing the benchmarking. However, in our case we are focused on evaluating the performance characteristics of the different models and their combinations. So, we benchmark the performance differences using identical data for all in order to make objective comparisons. We used the labeled Stanford Large Movie Review dataset (SLMRD) [1] in our experiments, because it is sufficiently large, well-curated, and challenging - see the discussion in Section .

Our experiments can be divided into two parts. The first part is the training of the individual participating models. The second part combines the participating models using our proposed sagittal stratum inspired neural network (SSNet) classifier. This allows for a clear and objective assessment of the advantages the proposed architecture offers compared to individual models.

The SLMRD dataset contains 25 000 labeled train reviews and 25 000 labeled test reviews.

We split the training data following the 5-folds cross-validation mechanism: (i) split the 25 000 train reviews into 5 folds; (ii) hold out each fold of 5 000 as validation data and use the remaining 20 000 reviews to train individual models. This generates 5 sets of predictions for each model. Then, we iterate through each set from all individual models and feed them as input to SSNet combiner to train it multiple times. In each trained classifier, we apply the entire test dataset with 25 000 reviews to measure its performance based on the accuracy of prediction. Finally, we use the mean average of these runs to report the skill of SSNet. By doing this, we hope we can compensate for the randomness brought by neural network models and provide a fair result.

Please note this split size (20K/5K) between training and validation is based on conventional best practices and is heuristic in nature. When designing the models and selecting the train dataset one needs to balance the number of trainable parameters in each model and the size of the training dataset to avoid under- and over-fitting.

We experimented with other split ratios using Train-Test split, e.g., 24K/1K, 23K/2K, 22K/3K, 21K/4K, 17.5K/7.5K, and the results were very similar to those presented below, hence we omitted these details. In all experiments presented in this section, we ensured that the models are trained properly.

## 3.2 Baseline Performance from Individual Models

The main idea behind the architecture in Section 2 is to incorporate separate views on the same corpus. In our experiments we used four models enumerated as $M_i$ $\forall i \in \{1, ..., 4\}$. In this section, we present an overview of the four models and their individual performance. We use 5-folds cross-validation to measure performance, i.e., using 20 000 reviews in training and 5 000 reviews in validation in each fold. Then, we provide results using the mean average of validation accuracy and the mean average of test accuracy on the entire 25 000 reviews in the test dataset.

### 3.2.1 BowTie

We use as $M_1$ the model described in [39]. It is based on the well-known bag-of-words model combined with word polarity. This model produces good results on the corpora [1, 21]. Here we introduced some minor tweaks to this model by incorporating few LSTM layers. This resulted in a small increase in the accuracy of the model.

We obtained a mean average of **89.55 %** validation accuracy and a mean average of **88.00 %** accuracy on the entire 25 000 reviews in the test dataset.

### 3.2.2 BLSTM with Attention and Glove Embeddings

In this model ($\boldsymbol{M_2}$), we use the glove embedding [32] on the dataset. While many other researchers in this area have obtained results by simply using LSTM or BLSTM with [32]; we found that the corpora [1, 21] contain reviews with nuances that are difficult to learn by simply passing the embeddings through LSTM or BLSTM. The models tend to learn the pattern of the inputs rather than the underlying meaning or semantics. This is often the cause of overfitting in a wide range of NLP problems. Further, in the case of sentiment analysis, certain words and their position in the sentence play extremely important role in determining the overall meaning. It is difficult to incorporate the positional semantics of these words using normal LSTM or BLSTM. The family of attention mechanisms [2, 16, 25] provides a direction to formulate such difficult semantics into the model. We revised the aforementioned attention mechanism to incorporate positional and sentimental semantics for sentences having large number of words.

Let $\overrightarrow{\boldsymbol{b}}$ and $\overleftarrow{\boldsymbol{b}}$ be the forward and the backward components of BLSTM and $\boldsymbol{k}$ be the sequence length, then $\boldsymbol{h} = [\overrightarrow{\boldsymbol{b}}, \overleftarrow{\boldsymbol{b}}]$ where $\boldsymbol{dim(h)} \in \mathbb{R}^{\boldsymbol{k} \times (|\overrightarrow{\boldsymbol{b}}| + |\overleftarrow{\boldsymbol{b}}|)}$. We define the following equations to describe the attention mechanism used in this paper:

$$\begin{aligned} h' &= tanh(h) \\ h' &= softmax(h') \\ C_v &= h \odot h' \\ M &= \sum_k C_v \end{aligned} \quad (16)$$

The third expression in the equation 16 represents the context vector $C_v$, which we sum up in the fourth expression over the sequence to remove stochastic behavior, hence $dim(M) \in \mathbb{R}^{(|\vec{b}|+|\overleftarrow{b}|)}$. To correctly calculate the *Hadamard product*, the vector space of $h'$ must be expanded after performing the *softmax* operation. This strategy inside the attention mechanism establishes a probabilistic vector space incorporating the positional and the sentimental semantics into $M_2$.

Our investigation showed that understanding nuances is not very computationally intensive but rather a logically inferential task, hence we used a low vector space, .i.e., 100 of the glove embeddings with sentence length equal to 800. This resulted in a small and effective model. We would also like to emphasize that the semantic structure of the language as understood by human brain is closely related to word embeddings (rather than language modeling). Hence, we did not incorporate any language modeling techniques in this architecture.

We obtained a mean average of **90.11 %** validation accuracy and a mean average of **89.78 %** accuracy on the entire 25 000 reviews in the test dataset.

### 3.2.3 BERT

BERT (Bidirectional Encoder Representations from Transformers) is the well-known large pre-trained model [14]. The BERT model is pre-trained on the minimally-filtered real-world text from Wikipedia (en.wikipedia.org) and BooksCorpus [41]. In this model ($M_3$), we fine-tuned BERT (https://tfhub.dev/tensorflow/bert_en_uncased_L-12_H-768_A-12/2) using SLMRD. The instance of BERT we used in our experiments has a maximum sequence length of 512. Please note that the average length of SLMRD is less than 300 words, but several go over 3000 words. This means that any text longer than 510 tokens (two required special tokens are added by BERT) gets truncated.

We obtained a mean average of **92.87 %** validation accuracy, and a mean average of **93.10 %** accuracy on the entire 25 000 reviews in the test dataset.

### 3.2.4 Universal Sentence Encoder (USE)

In this model ($M_4$), we experimented with the Universal Sentence Encoder [7] in embedding text. USE takes variable length text as input and encodes text into high-dimensional vectors that can be used for NLP tasks such as text classification, semantic similarity, etc. We use the highest allowed dimension of 512 in embedding text using USE (https://tfhub.dev/google/universal-sentence-encoder/4) in our training with SLMRD dataset.

We obtained a mean average of **88.98 %** validation accuracy and a mean average of **88.22 %** accuracy on the entire 25 000 reviews in the test dataset.

### 3.2.5 Performance Summary of Individual Models

Table 1 summarizes individual model performance based on the training on the split dataset with 20 000 train reviews and on the validation on the rest 5 000 train reviews.

The baseline performance is the test accuracy over the entire 25 000 reviews in the test dataset. $M_3$(BERT) achieves the best performance. Please note standard deviation (stdev) is followed after each mean accuracy across 5 folds.

**Table 1. Performance of Individual Models:**

| Individual Models Train on 20K | Mean Validation Accuracy (%) on 5K with stdev (SD) | Mean Accuracy (%) on Test (25K) with stdev (SD) |
|---|---|---|
| $M_1$ (BowTie) | 89.55 (0.245) | 88.00 (0.290) |
| $M_2$ (BLSTM) | 90.11 (0.68) | 89.78 (0.458) |
| $M_3$ (BERT) | 92.87 (0.407) | 93.10 (0.434) |
| $M_4$ (USE) | 88.98 (0.150) | 88.22 (0.155) |

### 3.3 Performance of Model Combiners

In this section we describe our experiments and present the computational results on proposed combiners. As discussed in previous sections, we have four models and three approaches to combine them. We evaluate each of the combiners on the basis of the accuracy achieved on the test dataset, consistent with the measurement criteria adopted for the performance of the individual models in Table 1. The combiner models train optimal weights $\{w_i\}_{i=1}^{4}$ for combining individual models using 20K/5K split in train dataset. First, get prediction probabilities of 5 000 train reviews from individual models as described in Section 3.2. Please note these individual models are trained using 20 000 train reviews. Each prediction probability value ranges from 0 to 1. Second, feed the 5 000 probability values to the combiner architecture as input and produce trained combining weights as the output. Finally, use the trained weights to combine the prediction probabilities of 25 000 test reviews produced from the same individual models. The combined probability value is mapped to positive sentiment if it is above **0.5**; otherwise it is mapped to negative sentiment. The results of the respective combiners are shown in Tables 2, 4, and 5. Similar to Section 3.2, we use mean test accuracy from multiple runs to report the results. Specifically, we repeated training the combiners **30** times using predictions in each fold generated from 5-folds cross-validation. Similar to Table 1, standard deviation (stdev) is shown after each mean accuracy value across all repeated runs.

#### 3.3.1 Neural Network Combiner

Table 2 shows the result obtained using the neural network combiner 2.1. Table 3 shows the trained combining weights. The neural network model is made up of a linear layer (without bias) followed by the sigmoid layer. Our combiner attained a maximum accuracy of **93.87 %**. The accuracy attained from the various individual model combinations varies but in all cases the combined model delivered higher prediction accuracy than any of the underlying individual models.

Figure 3 shows a graph of the train and test accuracy in addition to the test accuracy reported in Table 2 to illustrate that our models were trained appropriately.

One important element of our experiments is to confirm that the computed weights of the combined predictor in Section 2.1 conform to the estimate in 15. The coefficients computed for the combinations in Table 2 are shown in Table 3, and they are in agreement with the estimate 15. Depending on the optimizer configured in the computational environment [15] we used for these experiments, some coefficients subject to the constraint 2 may get set to zero if they go negative in any step of the training process. This is undesirable because the training process may not be able to recover well from such an event, and this would effectively disable the contribution of the corresponding contributing model in the combination. To improve the stability of the training process one may use different optimizers from the set of available options in their computational environment. For example, the environment [15] offers several good optimizers that can greatly reduce the occurrence of such events, e.g., the adaptive momentum (ADAM) optimizer, the Nesterov adaptive momentum (NADAM) optimizer, and the Root Mean Square Propagation (RMSPRop) optimizer. In addition, one may incorporate $L_2$-regularization in the dense layer predictor to improve the stability and increase the accuracy of the resulting combiner. We used ADAM with a $L_2$-weight of **0.039** for the computations in Tables 2 and 3.

**Table 2. Neural Network Combiner Accuracy:** A linear layer (without bias) and a sigmoid layer were used for this experiment. The maximum accuracy achieved is **93.87 %**.

| Combined Models | Mean Accuracy (%) on Test (25K) with stdev (SD) |
|---|---|
| $M_{1,2}$ | 90.80 (0.3201) |
| $M_{1,3}$ | 93.64 (0.2453) |
| $M_{1,4}$ | 89.89 (0.0847) |
| $M_{2,3}$ | 93.68 (0.1620) |
| $M_{2,4}$ | 91.14 (0.2337) |
| $M_{3,4}$ | 93.66 (0.2721) |
| $M_{1,2,3}$ | 93.76 (0.1966) |
| $M_{1,2,4}$ | 91.29 (0.2961) |
| $M_{1,3,4}$ | 93.82 (0.2652) |
| $M_{2,3,4}$ | 93.81 (0.2286) |
| $M_{1,2,3,4}$ | 93.87 (0.1832) |

**Table 3. Neural Network Combiner Mean Weights:**

| Combined Models | $w_{M_1}$ | $w_{M_2}$ | $w_{M_3}$ | $w_{M_4}$ |
|---|---|---|---|---|
| $M_{1,2}$ | 0.837 207 982 | 0.924 092 82 | - | - |
| $M_{1,3}$ | 0.862 401 351 | - | 1.218 124 133 | - |
| $M_{1,4}$ | 1.100 311 01 | - | - | 0.948 121 23 |
| $M_{2,3}$ | - | 0.843 419 33 | 1.291 315 655 | - |
| $M_{2,4}$ | - | 0.881 002 23 | - | 0.831 234 501 |
| $M_{3,4}$ | - | - | 1.128 714 311 | 0.599 008 786 |
| $M_{1,2,3}$ | 0.636 317 512 | 0.603 436 005 | 1.325 881 07 | - |
| $M_{1,2,4}$ | 0.799 691 307 | 0.971 208 091 | - | 0.779 012 99 |
| $M_{1,3,4}$ | 0.651 801 401 | - | 1.289 575 511 | 0.500 074 48 |
| $M_{2,3,4}$ | - | 0.533 645 1 | 1.141 277 1 | 0.540 118 14 |
| $M_{1,2,3,4}$ | 0.451 013 505 | 0.589 404 309 | 1.242 073 622 | 0.346 501 002 |

### 3.3.2 Bayesian Decision Rule Combiner

Here we present computational results from applying the Bayesian rules from [22]. The results in Table 4 are for the **Max**, **Avg**, **Sum**, and **Majority Vote** rules denoted as *max*, *avg*, *sum*, and *maj* correspondingly. We did not use the **Majority Vote** rule in the case of using two or four models, as is shown in Table 4, because of the potential tie in the vote. The different rules produced identical results when combining only two models but started to differentiate when the number of combined models increased.

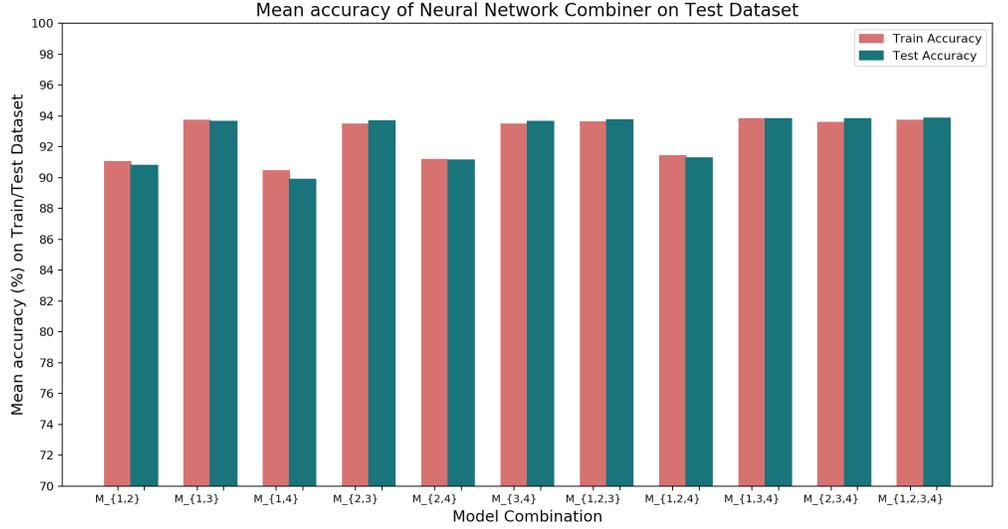

**Figure 3.** Mean train accuracy vs. Mean test accuracy in Neural Network Combiner. The plot shows the comparison of mean train accuracy and mean test accuracy for all the model combinations for the neural net combiner.

This is likely due to the low variance in the prediction probabilities with only a few models. Interestingly, in our case the **Max** rule tended to produce the highest accuracy unlike the results in [22] where the **Sum** rule was the best performer. This combined model performed well for the various combinations of individual models and rules.

### 3.3.3 Heuristic-Hybrid Combiner

Table 5 shows the results obtained using the heuristic-hybrid combiner. Each of the individual models is used as base, and when its prediction probability falls below $\theta$, where $0.5 < \theta < 1$, then combinations of the auxiliary models are used for the prediction. We use the Bayesian decision rules from [22] to compute a prediction with the auxiliary models. The maximum accuracy achieved on the test dataset is **93.73 %** with model $M_3$ as base and models $M_1$, $M_2$ and $M_4$ as auxiliary. The experimental data in Table 5 bear out the recommendation from Section 2.3 to use the best model as base. The results for $M_3$ as base show a good balance of collaboration with the auxiliary models with $\theta = 0.91$ to deliver the best accuracy result. Again, the accuracy obtained by the $max$ rule tended to perform best. The maximum accuracy attained by this combiner was higher than that of the Bayesian Decision Rule combiners from [22], cf. Table 4, and in this sense our heuristic-hybrid combiner performed better.

**Table 4. Accuracy of the Bayesian Decision Rule Combiner:** The indication **all** for the rule used means that all rules except ***maj*** produced the same result. The ***maj*** rule was **not** used in the case of combining an even number of models to avoid a tie in the vote. The maximum accuracy achieved is **93.63 %**.

| Combined Models | Rule Used | Mean Accuracy (%) on Test (25K) with stdev (SD) |
|---|---|---|
| $M_{1,2}$ | all | 90.69 (0.4052) |
| $M_{1,3}$ | all | 93.46 (0.4320) |
| $M_{1,4}$ | all | 89.86 (0.0585) |
| $M_{2,3}$ | all | 93.56 (0.2450) |
| $M_{2,4}$ | all | 91.03 (0.1974) |
| $M_{3,4}$ | all | 93.52 (0.2304) |
| $M_{1,2,3}$ | max | 93.58 (0.2829) |
| | avg | 92.93 (0.3248) |
| | sum | 92.93 (0.3248) |
| | maj | 92.39 (0.2828) |
| $M_{1,2,4}$ | max | 91.16 (0.1707) |
| | avg | 91.12 (0.2860) |
| | sum | 91.12 (0.2860) |
| | maj | 90.71 (0.2687) |
| $M_{1,3,4}$ | max | 93.50 (0.2657) |
| | avg | 92.46 (0.2237) |
| | sum | 92.46 (0.2237) |
| | maj | 91.94 (0.1597) |
| $M_{2,3,4}$ | max | 93.63 (0.2326) |
| | avg | 93.06 (0.2277) |
| | sum | 93.06 (0.2277) |
| | maj | 92.67 (0.1875) |
| $M_{1,2,3,4}$ | max | 93.60 (0.2443) |
| | avg | 92.89 (0.2154) |
| | sum | 92.89 (0.2154) |

# 4 Conclusions and next steps

We successfully followed our intuition inspired by the biological underpinning of the human brain for understanding sarcasm to construct a neural network architecture for sentiment analysis. We considered novel systematic and heuristic-hybrid implementations of the framework and were able to provide theoretical justification for the state-of-the-art computational performance of the best systematic solution based on the neural network dense layer. Our heuristic-hybrid solution closely followed the intuition inspired by the biological underpinnings of our brain while at the same time relied on a systematic technique for combining the predictions of the auxiliary models using well-known Bayesian rules. This approach delivered performance results very close to those of the best combined predictor. Thus, our two novel combined models outperformed not only each individual auxiliary model in terms of accuracy and robustness but also the legacy combiner models from the literature. A graphical comparison of the three approaches is shown in Figure 4.

The combiners we considered are built on the notion that if the combined models are sufficiently decorrelated then the combination can mitigate some of the shortcomings of the individual models. Our approach to combining models and the theoretical result for **Figure 4** shows the best attained accuracy of individual participating models and their combinations for the proposed approach.

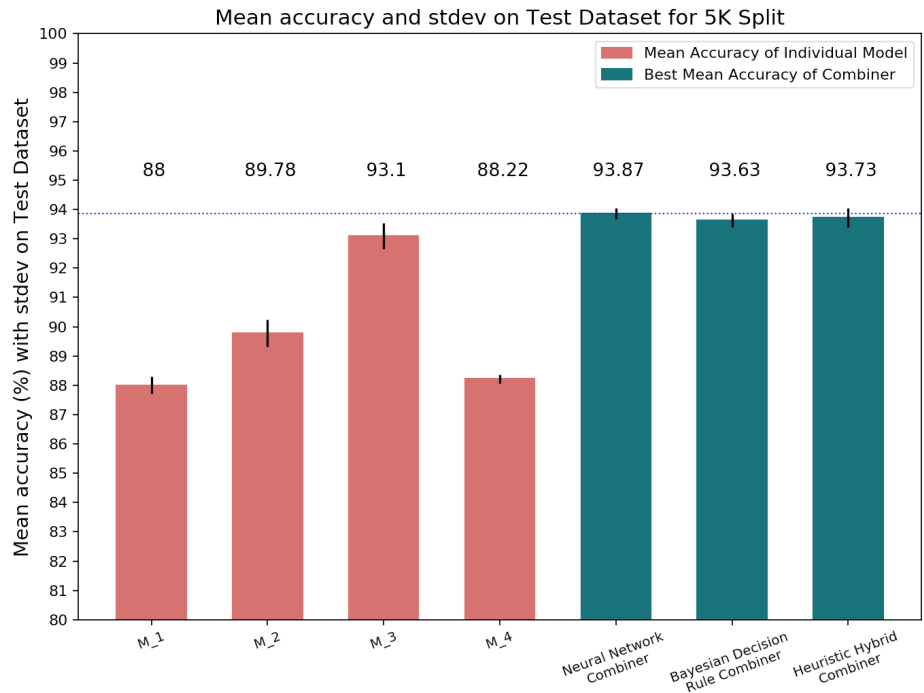

**Figure 4.** Accuracy on test dataset for proposed methods

the systemic combiner are generic and may be used in other classifier application domains. Moreover, in our framework each individual model is a pluggable component, and this opens up the possibility to combine diverse multi-modal inputs to solve various machine learning tasks and improve the robustness against adversarial attacks tailored for specific models. We plan to investigate these aspects in the future.

## Acknowledgments

We thank the NIST Information Technology Laboratory (ITL) for the research funding and support.

**Table 5. Heuristic-Hybrid Combiner:** Each model was used as a base model while $max$, $avg$, $sum$, and $maj$ rules were used for predicting with the auxiliary models when the confidence of the base model fell below the threshold $\theta$. The maximum accuracy of **93.73 %** was attained with $\boldsymbol{M_3}$ as the base model for the rule: $\boldsymbol{max}$.

| Combined Models | | | | Mean Accuracy (%) on Test (25K) with stdev (SD) |
|---|---|---|---|---|
| Base | Auxiliary | Rule Used | Mean Threshold $\theta$ | |
| $M_1$ | $M_2$ | - | 0.91 | 90.19 (0.3709) |
| | $M_3$ | - | 0.97 | 93.09 (0.4306) |
| | $M_4$ | - | 0.85 | 89.49 (0.1617) |
| | $M_{2,3}$ | max | 0.99 | 93.49 (0.3107) |
| | | avg | 0.99 | 93.49 (0.3107) |
| | | sum | 0.99 | 93.49 (0.3107) |
| | $M_{2,4}$ | max | 0.96 | 91.06 (0.1698) |
| | | avg | 0.96 | 91.06 (0.1698) |
| | | sum | 0.96 | 91.06 (0.1698) |
| | $M_{3,4}$ | max | 0.99 | 93.45 (0.2848) |
| | | avg | 0.99 | 93.45 (0.2848) |
| | | sum | 0.99 | 93.45 (0.2848) |
| | $M_{2,3,4}$ | max | 0.99 | 93.56 (0.2872) |
| | | avg | 0.99 | 93.00 (0.2462) |
| | | sum | 0.99 | 93.00 (0.2462) |
| | | maj | 0.98 | 92.59 (0.1596) |
| $M_2$ | $M_1$ | - | 0.82 | 90.49 (0.3983) |
| | $M_3$ | - | 0.98 | 93.28 (0.3105) |
| | $M_4$ | - | 0.81 | 90.76 (0.3105) |
| | $M_{1,3}$ | max | 0.99 | 93.47 (0.3638) |
| | | avg | 0.99 | 93.47 (0.3638) |
| | | sum | 0.99 | 93.47 (0.3638) |
| | $M_{1,4}$ | max | 0.88 | 91.02 (0.2680) |
| | | avg | 0.88 | 91.02 (0.2680) |
| | | sum | 0.88 | 91.02 (0.2680) |
| | $M_{3,4}$ | max | 0.98 | 93.53 (0.2389) |
| | | avg | 0.98 | 93.53 (0.2389) |
| | | sum | 0.98 | 93.53 (0.2389) |
| | $M_{1,3,4}$ | max | 0.99 | 93.50 (0.2535) |
| | | avg | 0.97 | 92.57 (0.1870) |
| | | sum | 0.97 | 92.57 (0.1870) |
| | | maj | 0.97 | 92.18 (0.1497) |
| $M_3$ | $M_1$ | - | 0.92 | 93.36 (0.3938) |
| | $M_2$ | - | 0.85 | 93.47 (0.2438) |
| | $M_4$ | - | 0.90 | 93.50 (0.2670) |
| | $M_{1,2}$ | max | 0.91 | 93.58 (0.3228) |
| | | avg | 0.91 | 93.58 (0.3228) |
| | | sum | 0.91 | 93.58 (0.3228) |
| | $M_{1,4}$ | max | 0.91 | 93.60 (0.3346) |
| | | avg | 0.91 | 93.60 (0.3346) |
| | | sum | 0.91 | 93.60 (0.3346) |
| | $M_{2,4}$ | max | 0.91 | 93.71 (0.3039) |
| | | avg | 0.91 | 93.71 (0.3039) |
| | | sum | 0.91 | 93.71 (0.3039) |
| | $\boldsymbol{M_{1,2,4}}$ | **max** | **0.91** | **93.73 (0.3259)** |
| | | avg | 0.91 | 93.68 (0.3320) |
| | | sum | 0.91 | 93.68 (0.3320) |
| | | maj | 0.91 | 93.62 (0.3141) |
| $M_4$ | $M_1$ | - | 0.89 | 89.55 (0.1175) |
| | $M_2$ | - | 0.96 | 90.72 (0.2550) |
| | $M_3$ | - | 0.99 | 93.20 (0.2550) |
| | $M_{1,2}$ | max | 0.97 | 90.98 (0.3637) |
| | | avg | 0.97 | 90.98 (0.3637) |
| | | sum | 0.97 | 90.98 (0.3637) |
| | $M_{1,3}$ | max | 0.99 | 93.30 (0.3262) |
| | | avg | 0.99 | 93.30 (0.3262) |
| | | sum | 0.99 | 93.30 (0.3262) |
| | $M_{2,3}$ | max | 0.99 | 93.30 (0.1960) |
| | | avg | 0.99 | 93.42 (0.1960) |
| | | sum | 0.99 | 93.42 (0.1960) |
| | $M_{1,2,3}$ | max | 0.99 | 93.42 (0.2397) |
| | | avg | 0.99 | 92.82 (0.3129) |
| | | sum | 0.99 | 92.82 (0.3129) |
| | | maj | 0.99 | 92.36 (0.2764) |